\documentclass[letterpaper]{article} 
\usepackage{aaai2027}  
\usepackage[hyphens]{url}  
\usepackage{graphicx} 
\usepackage{natbib}  
\usepackage{caption} 
\usepackage{algorithm}
\usepackage{algorithmic}

\usepackage{newfloat}
\usepackage{listings}
\DeclareCaptionStyle{ruled}{labelfont=normalfont,labelsep=colon,strut=off} 
\floatstyle{ruled}
\newfloat{listing}{tb}{lst}{}
\floatname{listing}{Listing}

\usepackage{booktabs}
\usepackage{amsmath}
\usepackage{amssymb}
\usepackage{mathtools}
\usepackage{graphicx}
\usepackage{multirow}
\usepackage{subcaption}
\title{GUPO: Gradient Uncertainty-aware Policy Optimization for Post-Training Large Language Models}
\copyrighttext{Preprint. Under Review.}
\author{
    Peizheng Guo\textsuperscript{\rm 1,2,3}\equalcontrib,
    Jianqi Zhang\textsuperscript{\rm 1,2,3}\equalcontrib,
    Xingyu Zhang\textsuperscript{\rm 1,2,3},
    Yun Fan\textsuperscript{\rm 3},
    Jiahuan Zhou\textsuperscript{\rm 4},\\
    Changwen Zheng\textsuperscript{\rm 1,2,3},
    Wenwen Qiang\textsuperscript{\rm 1,2,3}\corresponding
}
\affiliations{
    \textsuperscript{\rm 1}Institute of Software Chinese Academy of Sciences,
    \textsuperscript{\rm 2}University of Chinese Academy of Sciences,\\
    \textsuperscript{\rm 3}National Key Laboratory of Space Integrated Information System,\\
    \textsuperscript{\rm 4}Wangxuan Institute of Computer Technology, Peking University

}

\begin{document}

\maketitle

\begin{abstract}
Group Relative Policy Optimization (GRPO) has become a widely used approach for post-training Large Language Models (LLMs) for reasoning. In GRPO, the group gradients induced by different queries within the same mini-batch are directly averaged to form the policy update. However, these group gradients can point in conflicting directions. Our empirical analysis suggests that group-gradient conflicts tend to be associated with less effective policy updates, motivating the need for a reliable aggregated update direction under such conflicts. Standard GRPO aggregation treats the realized group gradients as deterministic contributions and does not account for differences in their reliability during aggregation. To address this issue, we propose Gradient Uncertainty-Aware Policy Optimization (GUPO), which models each group gradient as a random variable under a Bayesian formulation and estimates its probability distribution. GUPO then derives gradient uncertainty using a Dirichlet-based formulation and uses it to calibrate the contribution of each group gradient during aggregation. Extensive experiments on multiple benchmarks demonstrate the effectiveness of GUPO.
\end{abstract}

\section{Introduction}
\label{sec:introduction}
Large language models (LLMs) have demonstrated strong capabilities in language understanding, text generation, and complex reasoning
\cite{brown2020language,chowdhery2023palm,kojima2022large,wei2022chain}.
Reinforcement-learning-based post-training, particularly Group Relative Policy Optimization (GRPO) \cite{shao2024deepseekmath}, has become an important approach for further improving the reasoning capabilities of LLMs.
For each query, GRPO samples a group of responses, derives group-relative advantages from their rewards, and uses these advantages to weight token-level policy-gradient terms.
The resulting gradient contributions are aggregated to update the policy parameters.

\begin{figure}[t]
    \centering
    \includegraphics[width=\columnwidth]{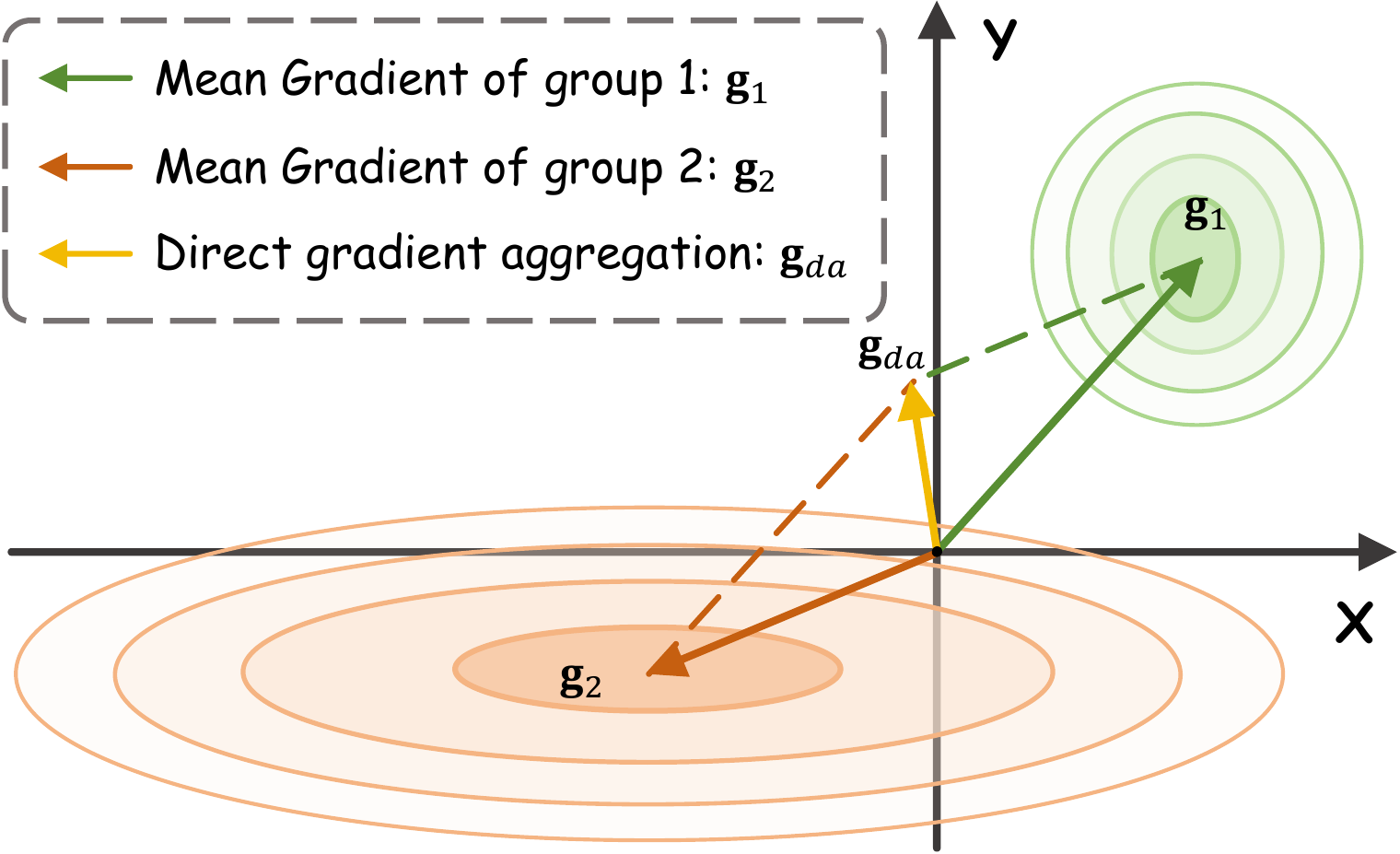}
    \caption{Explanation of gradient uncertainty and conflicts. Two-dimensional example of two group gradients from the same GRPO mini-batch.}
    \label{fig:conflict_illustration}
\end{figure}

\begin{figure*}[t]
    \centering
    \begin{subfigure}[t]{0.46\columnwidth}
        \centering
        \includegraphics[width=\linewidth]{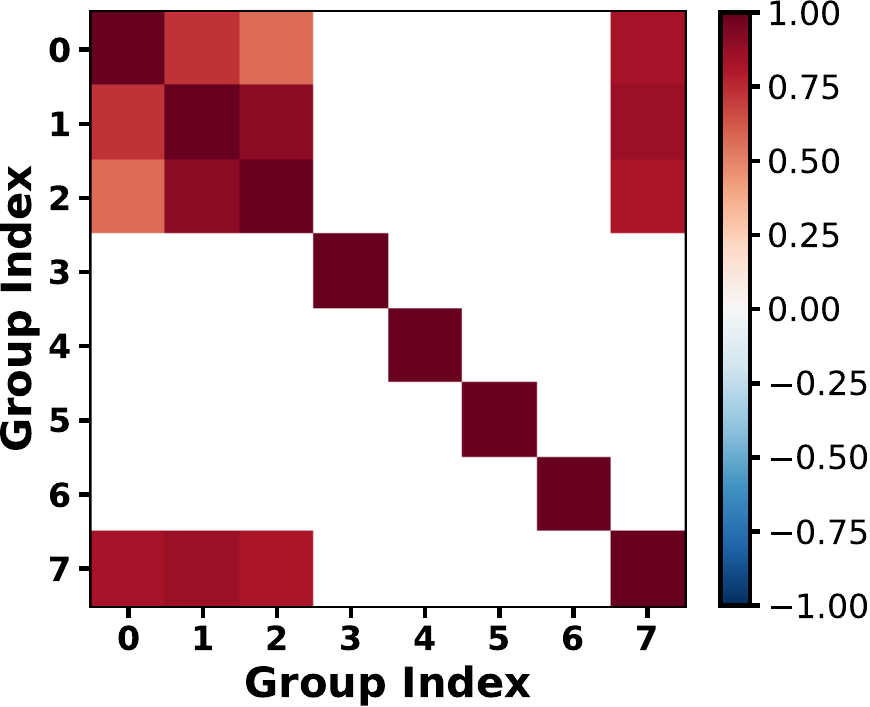}
        \caption{Low-Conflict Mini-batch}
        \label{fig:motivation_low}
    \end{subfigure}
    \hfill
    \begin{subfigure}[t]{0.46\columnwidth}
        \centering
        \includegraphics[width=\linewidth]{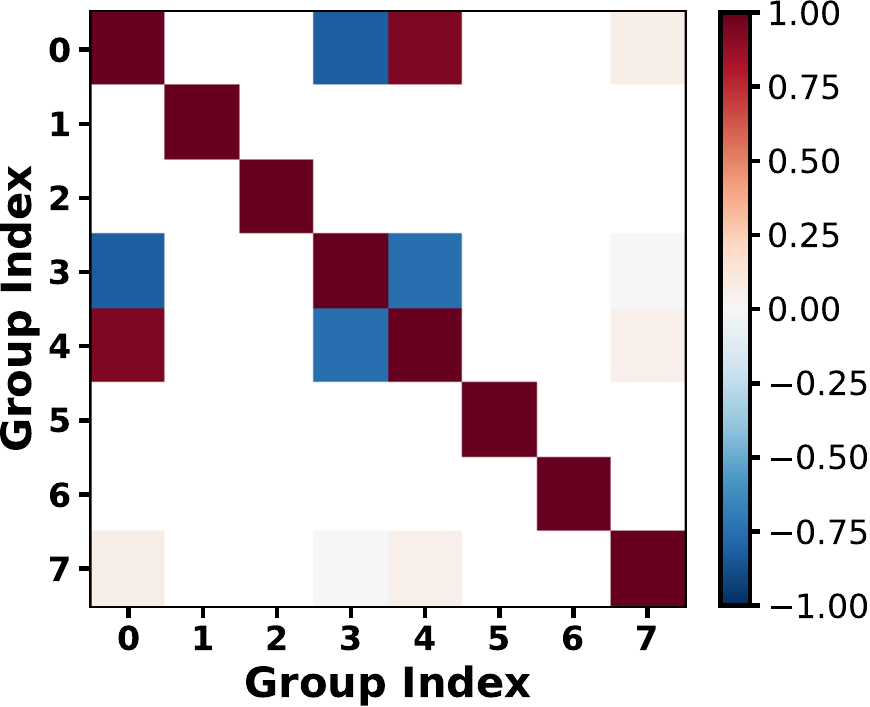}
        \caption{High-Conflict Mini-batch}
        \label{fig:motivation_high}
    \end{subfigure}
    \hfill
    \begin{subfigure}[t]{0.54\columnwidth}
        \centering
        \includegraphics[width=\linewidth]{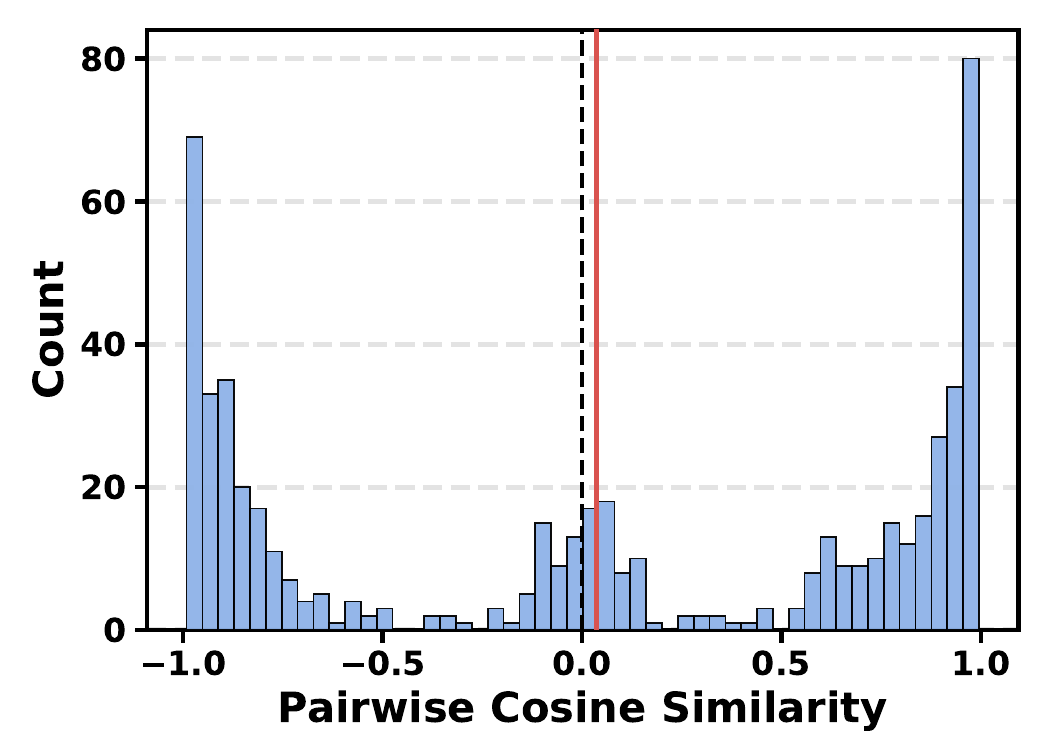}
        \caption{Conflict Distribution}
        \label{fig:motivation_pair}
    \end{subfigure}
    \hfill
    \begin{subfigure}[t]{0.62\columnwidth}
        \centering
        \includegraphics[width=\linewidth]{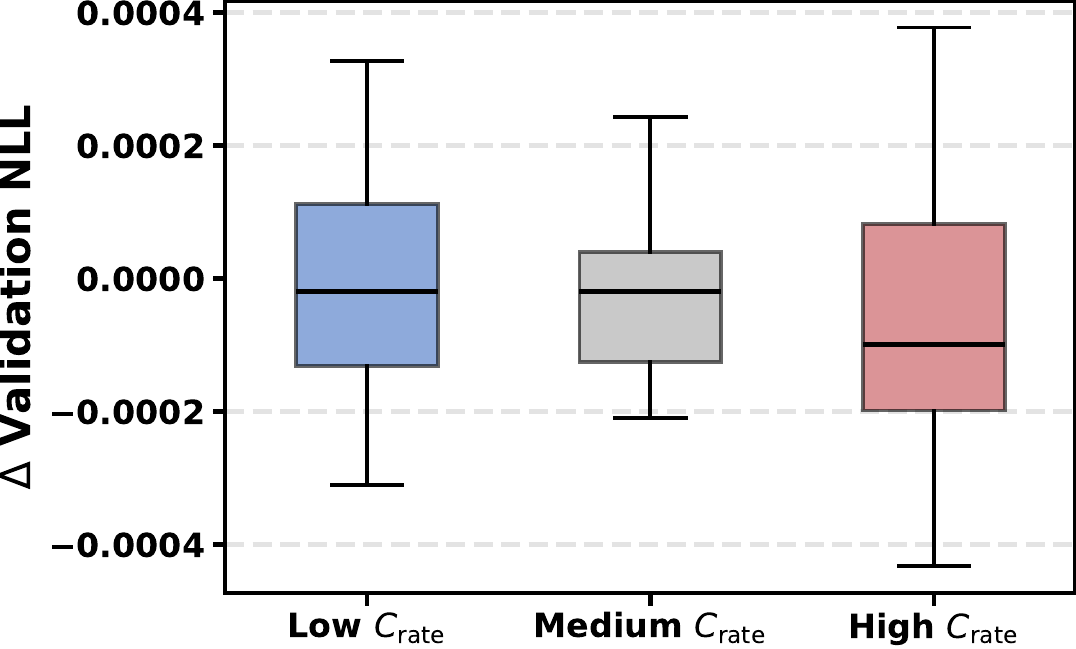}
        \caption{Conflict v.s. Performance}
        \label{fig:motivation_bin}
    \end{subfigure}
    \caption{Empirical results of group-gradient conflicts in GRPO. (a) and (b) show the pairwise cosine-similarity matrices of representative mini-batches with low and high conflict levels. (c) presents the distribution of pairwise cosine similarities among group gradients. (d) compares the validation $\Delta\mathrm{NLL}$ across mini-batches with low, medium, and high conflict rates.}
    \label{fig:motivation}
\end{figure*}

\begin{figure*}[t]
    \centering
    \includegraphics[width=\textwidth]{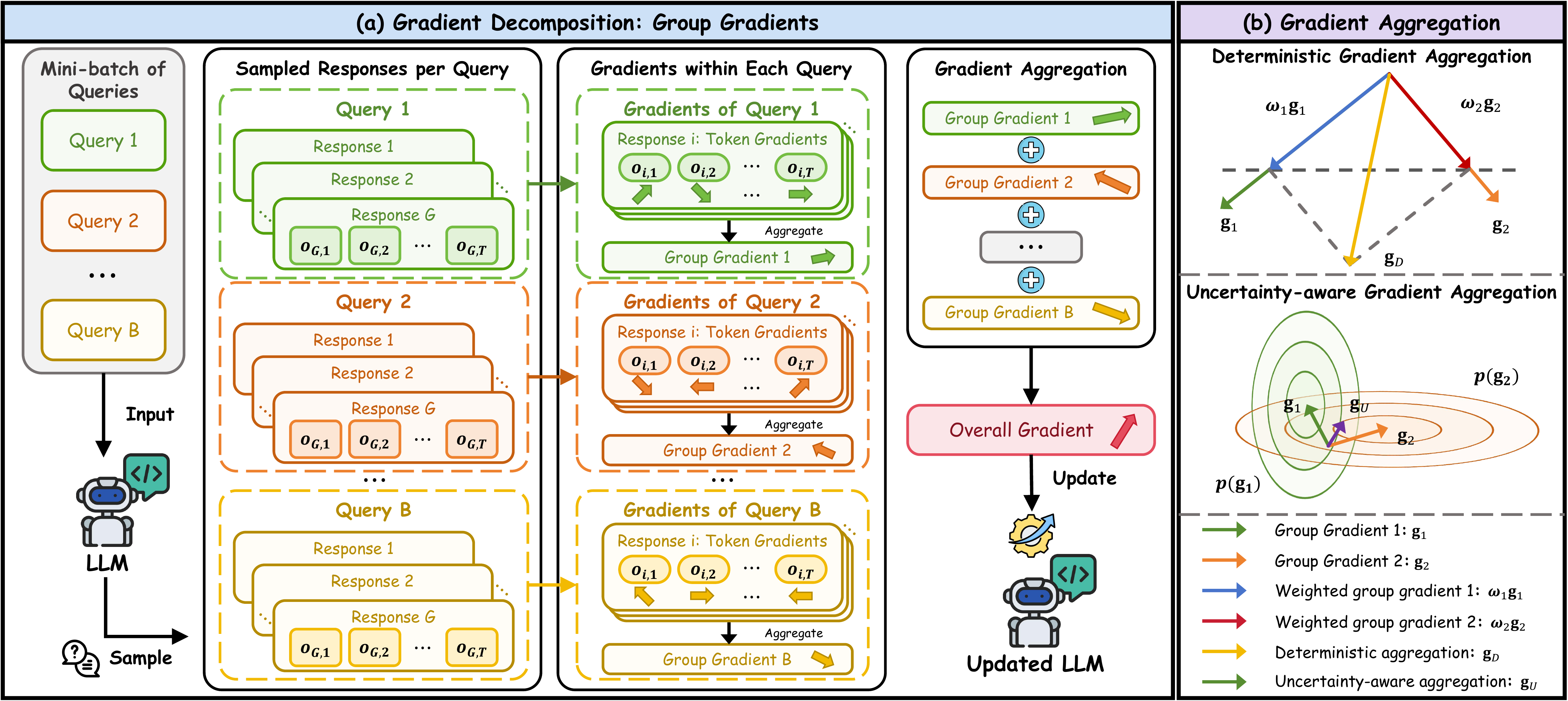}
    \caption{Illustration of GRPO gradient optimization process.}
    \label{fig:motivation_illustration}
\end{figure*}

In GRPO, for each query, the gradients induced by its sampled response group are aggregated into a \textit{group gradient}, which represents the contribution of that query group to the mini-batch update.
The mini-batch gradient is then obtained by averaging the group gradients across queries, as illustrated in \textbf{Figure~\ref{fig:motivation_illustration}} \cite{ge2026grpo,panaganti2026group,nguyen2026adaptive,barakat2026pass}.
Based on this update mechanism, we observe an interesting phenomenon: group gradients induced by different queries within the same mini-batch may point in conflicting directions. 
Specifically, for each mini-batch, we compute the group gradients and quantify their pairwise directional conflicts. We then analyze the association between the gradient conflict and the effectiveness of the corresponding policy update.
As shown in \textbf{Figure \ref{fig:motivation}}, group gradients are largely aligned in some mini-batches, whereas clear directional conflicts emerge in others. Mini batches with severe conflicts tend to exhibit limited validation update gains.
They motivate the central question of this work:
\textbf{when group gradients induced by different queries conflict within the same GRPO mini-batch, how can we obtain a reliable aggregated update direction?}

In this context, a reliable update direction can be understood as one that reasonably reflects the overall optimization tendency supported by the mini-batch, rather than being overly influenced by a few conflicting gradients. 
Gradient uncertainty provides a possible basis for assessing such reliability of different group gradients during aggregation \cite{ghavamzadeh2016bayesian,akella2021deep}. 
Specifically, group gradients can be treated as random variables and represented as probability distributions that describe the range of possible gradient values and their likelihoods.
If the distribution is concentrated, most of its probability mass lies within a narrow range around the mean, indicating that the gradient values supported by the distribution are similar.
Such limited variation among these values corresponds to lower gradient uncertainty and higher reliability.
If the distribution is more dispersed, its probability mass extends over a broader range of substantially different gradient values.
This larger variation among these values corresponds to higher gradient uncertainty and lower reliability.
Figure~\ref{fig:conflict_illustration} presents a two-dimensional example for illustration.
$\mathbf{g}_1$ and $\mathbf{g}_2$ represent the mean gradients of two query groups, while the surrounding contours represent their probability distributions.
The distribution of $\mathbf{g}_1$ concentrates more probability mass near the mean gradient, corresponding to lower uncertainty and higher reliability.
By contrast, the distribution of $\mathbf{g}_2$ covers a broader range of gradient values, corresponding to higher uncertainty and lower reliability.
Direct aggregation assigns the two group gradients equal coefficients without accounting for this difference in reliability.
Consequently, the more uncertain group gradient may exert excessive influence on the directly aggregated direction $\mathbf{g}_{\mathrm{da}}$.
Based on this analysis, we propose to explicitly model group-gradient uncertainty and incorporate it into gradient aggregation, emphasizing more reliable group gradients while attenuating the influence of less reliable ones, with the aim of obtaining a reliable update direction when gradient conflicts occur.

To this end, we propose Gradient Uncertainty-aware Policy Optimization (GUPO), explicitly modeling the uncertainty of group gradients within each mini-batch to achieve reliable policy updates.
Rather than treating each group gradient solely as a deterministic vector, we model it as a random variable represented by a probability distribution.
Specifically, under a Bayesian formulation \cite{dempster1968generalization,box2011bayesian}, we first estimate the probability distribution of each group gradient. The precision of the resulting distribution is then mapped to gradient evidence. Following the relationship among evidence, belief, and uncertainty in subjective logic \cite{jsang2018subjective,bao2021evidential,sensoy2018evidential}, we derive dimension-wise beliefs and the overall uncertainty of each group gradient through a Dirichlet-based evidential formulation. 
The resulting group-level uncertainty is incorporated into gradient aggregation, increasing the relative contribution of lower-uncertainty group gradients while attenuating that of higher-uncertainty ones. In this way, GUPO aims to mitigate the excessive influence of highly uncertain gradients and obtain a more reliable aggregated update direction when group-gradient conflicts occur.
Experiments across multiple benchmarks suggest the effectiveness of our method.

The main contributions can be summarized as follows:
(i) We identify directional conflicts among group gradients induced by different queries within the same GRPO mini-batch, and empirically show that conflicts can be associated with less effective policy updates, motivating the problem of obtaining a reliable aggregated update direction.
(ii) We analyze this problem from the perspective of gradient uncertainty and propose GUPO, which models each group gradient as a probability distribution under a Bayesian formulation, derives its uncertainty through a Dirichlet-based evidential formulation, and incorporates the resulting uncertainty into gradient aggregation.
(iii) Extensive experiments across multiple benchmarks demonstrate the effectiveness of GUPO, showing that incorporating group-gradient uncertainty into policy optimization can improve GRPO-based post-training.

\section{Related Work}
\label{sec:related_work}
\subsubsection{Gradient Conflict in LLM Post-Training}
\label{sec:rw_gc_llm}
Reinforcement learning has become an important approach for LLM post-training, with GRPO \cite{shao2024deepseekmath} being widely studied for improving reasoning capabilities.
However, different gradients produced during GRPO optimization may result in incompatible directions, thereby interfering with effective policy updates. 
Recent studies investigate such gradient conflicts in GRPO-based post-training at different levels of the optimization process.
DaGRPO \cite{xie2025dagrpo} investigates conflicts within a rollout group, relating them to insufficient distinctiveness among sampled responses and applying sequence-level gradient correction.
PCR \cite{qiang2026plasticity} studies conflicts between plasticity and stability gradients in GRPO and addresses them through probabilistic projection.
ResRL \cite{lin2026resrl} analyzes gradient interference between positive and negative responses and modulates negative gradients using projection residuals.
These studies indicate that gradient conflicts can arise from different sources and at different granularities.
Unlike prior works, we focus on conflicts among group gradients associated with different queries within the same GRPO mini-batch, and analyze them from the perspective of gradient uncertainty. We propose an optimization framework that explicitly characterizes the uncertainty of these query-level group gradients and calibrates their contributions during mini-batch gradient aggregation.

\subsubsection{Uncertainty Estimation in LLM Post-Training}
\label{sec:rw_uncertainty}
Uncertainty estimation provides a principled way to quantify the reliability of model outputs or evaluation signals and is widely applied to LLMs to assess the reliability of generated responses and alignment signals.
Uncertainty-aware Reward Model \cite{lou2024uncertainty} quantifies uncertainty in learned reward-model predictions by modeling stochastic human preferences and disagreement among reward models.
SEED-GRPO \cite{chen2025seed} estimates query-level uncertainty from the semantic diversity of multiple sampled responses and uses it to modulate the corresponding advantages.
CAPO \cite{wang2026calibration} studies the overconfidence associated with uncertainty-agnostic advantage estimation in GRPO and introduces uncertainty-aware advantages to improve reasoning performance and confidence calibration.
Recent works show that uncertainty can serve as an explicit measure of how reliably a prediction or scalar evaluation should be trusted. 
Unlike previous methods, we focus on conflicts among group gradients associated with different queries within the same GRPO mini-batch and analyze them from the perspective of gradient uncertainty. We model each group gradient as a random variable represented by a probability distribution under a Bayesian formulation, and derive an uncertainty estimation to reflect the reliability of its update signal. The resulting uncertainty is further incorporated into gradient aggregation to mitigate the adverse effects of gradient conflicts and guide more effective policy updates.

\section{Problem Settings and Analyses}
\label{sec:problem_setting_analyses}

\subsubsection{Problem Settings}
\label{sec:problem_setting}
For a given query $q \sim P(Q)$, GRPO samples a group of $G$ candidate outputs $\{o_i\}_{i=1}^{G}$ from the old policy $\pi_{\theta_{\mathrm{old}}}$. Each output $o_i$ is generated autoregressively as $o_i = \{o_{i,1}, o_{i,2}, \dots, o_{i,T_i}\}$, where $T_i$ denotes the length of the $i$-th output. After generation, each output is evaluated by an outcome reward function $r_i = R_{\mathrm{out}}(q,o_i)$. In verifiable reasoning tasks, $R_{\mathrm{out}}$ is usually defined according to whether the final answer is correct. 

GRPO estimates the relative advantage of each output within the sampled group, encouraging the policy to increase the probability of outputs that obtain higher rewards.
\begin{equation}
A_i =
\frac{
r_i-\mathrm{mean}(r_1,r_2,\dots,r_G)
}{
\mathrm{std}(r_1,r_2,\dots,r_G)
}.
\end{equation}

The GRPO objective can then be written as
\begin{equation}
\begin{aligned}
&\mathcal{J}_{\mathrm{GRPO}}(\theta)
=\mathbb{E}_{q\sim P(Q),\{o_i\}_{i=1}^{G}\sim\pi_{\theta_{\mathrm{old}}}}\\
&\scalebox{0.78}{$\displaystyle
\frac{1}{G}\sum_{i=1}^{G}\frac{1}{T_i}\sum_{t=1}^{T_i}\bigg[\min (\rho_{i,t}(\theta)A_i,\mathrm{clip}(\rho_{i,t}(\theta),1-\varepsilon,1+\varepsilon)A_i)-\beta D_{\mathrm{KL}}^{i,t}\bigg]
$},
\end{aligned}
\end{equation}
where $\varepsilon$ is the clipping coefficient and $\beta$ controls the strength of KL regularization.
GRPO uses the importance ratio between the current policy $\pi_\theta$ and the old policy $\pi_{\theta_{\mathrm{old}}}$:
\begin{equation}
\rho_{i,t}(\theta)
=
\frac{
\pi_{\theta}(o_{i,t}\mid q,o_{i,<t})
}{
\pi_{\theta_{\mathrm{old}}}(o_{i,t}\mid q,o_{i,<t})
}.
\end{equation}

The KL regularization is computed with respect to a fixed reference policy $\pi_{\mathrm{ref}}$, which can be estimated as
\begin{equation}
D_{\mathrm{KL}}^{i,t}
=
\frac{
\pi_{\mathrm{ref}}(o_{i,t}\mid q,o_{i,<t})
}{
\pi_{\theta}(o_{i,t}\mid q,o_{i,<t})
}
-
\log
\frac{
\pi_{\mathrm{ref}}(o_{i,t}\mid q,o_{i,<t})
}{
\pi_{\theta}(o_{i,t}\mid q,o_{i,<t})
}
-1.
\end{equation}

For an optimization mini-batch containing $B$ queries, we focus on the advantage-weighted policy-gradient signals that determine how the sampled responses contribute to the policy update. The corresponding mini-batch gradient is
\begin{equation}
\begin{aligned}
    \nabla_{\theta}\mathcal{J}_{\mathcal{B}}(\theta)=\frac{1}{BG}&\sum_{b=1}^{B}\sum_{i=1}^{G}\frac{1}{T_{b,i}}\sum_{t=1}^{T_{b,i}}\\
    A_{b,i}\rho_{b,i,t}(\theta)\nabla_{\theta}\log&\pi_{\theta}\left(o_{b,i,t}\mid q_b,o_{b,i,<t}\right).
\end{aligned}
\label{eq:grpo_minibatch_gradient}
\end{equation}

This gradient can be decomposed according to the query groups within the mini-batch \cite{ge2026grpo,panaganti2026group,nguyen2026adaptive,barakat2026pass}.
Specifically, the gradient contributed by the $b$-th query is defined as
\begin{equation}
\begin{aligned}
    \mathbf{g}_b(\theta)=\frac{1}{G}\sum_{i=1}^{G}&\frac{1}{T_{b,i}}\sum_{t=1}^{T_{b,i}}\\
    A_{b,i}\rho_{b,i,t}(\theta)\nabla_{\theta}\log\pi_{\theta}&\left(o_{b,i,t}\mid q_b,o_{b,i,<t}\right),
\end{aligned}
\label{eq:group_gradient}
\end{equation}
which we term the \emph{group gradient}. Accordingly, the overall gradient can be expressed as
\begin{equation}
\nabla_{\theta}\mathcal{J}_{\mathcal{B}}(\theta)=\frac{1}{B}\sum_{b=1}^{B}\mathbf{g}_b(\theta).
\label{eq:group_gradient_decomposition}
\end{equation}

\begin{figure*}[t]
    \centering
    \includegraphics[width=\textwidth]{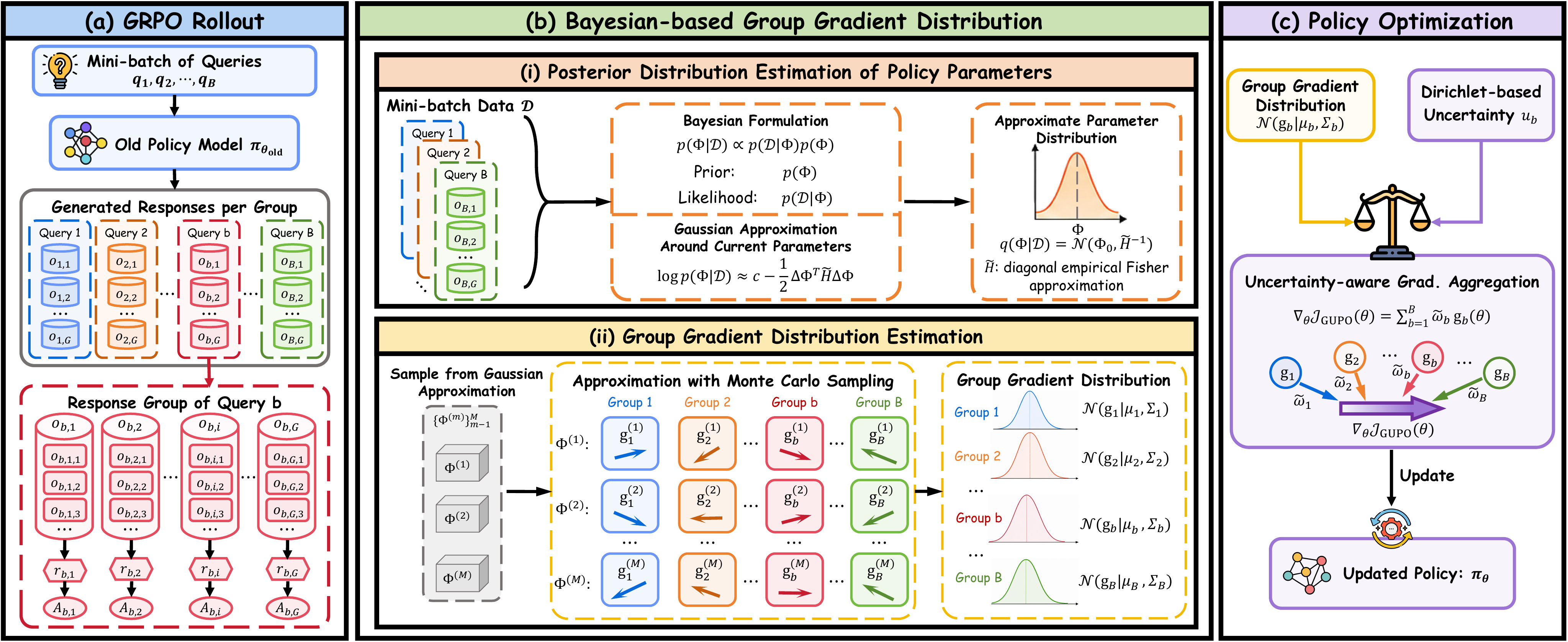}
    \caption{Overview of our proposed method.}
    \label{fig:overview}
\end{figure*}

\subsubsection{Empirical Analysis}
\label{sec:empirical_analysis}
In GRPO, the policy gradient of an optimization mini-batch can be decomposed as the aggregation of group gradients from different queries \cite{ge2026grpo,panaganti2026group,nguyen2026adaptive,barakat2026pass}. As illustrated in \textbf{Figure~\ref{fig:motivation_illustration}a}, the token-level gradients from the $G$ responses for query $q_b$ are aggregated into the group gradient $\mathbf{g}_b(\theta)$. For a mini-batch containing $B$ queries, the gradient is obtained by averaging these group gradients: $\nabla_{\theta}\mathcal{J}_{\mathcal{B}}(\theta)=\frac{1}{B}\sum_{b=1}^{B}\mathbf{g}_b(\theta)$.

Based on this update mechanism, we conduct controlled experiments. Starting from the same checkpoint, we sample different mini-batches under identical settings, each containing $B$ query groups with $G$ responses per query. 
For each group-gradient pair, we compute the cosine similarity $c_{b,b'}=\frac{\mathbf{g}_b^{\top}\mathbf{g}_{b'}}{\lVert\mathbf{g}_b\rVert_2\lVert\mathbf{g}_{b'}\rVert_2}$, where $c_{b,b'}<0$ indicates a directional conflict. The conflict level of mini-batch $\mathcal{B}$ is measured by $C_{\mathrm{rate}}(\mathcal{B})=\frac{\sum_{(b,b')\in\mathcal{P}_{\mathcal{B}}}\mathbb{I}[c_{b,b'}<0]}{|\mathcal{P}_{\mathcal{B}}|}$, where $\mathcal{P}_{\mathcal{B}}$ is the set of group-gradient pairs and $\mathbb{I}[\cdot]$ is the indicator function. To evaluate the corresponding policy update, we perform GRPO update using each mini-batch and compute $\Delta\mathrm{NLL}=\mathrm{NLL}_{\mathrm{before}}-\mathrm{NLL}_{\mathrm{after}}$ on a fixed validation set, where a larger $\Delta\mathrm{NLL}$ indicates a more effective update. More details are provided in the \textbf{Appendix}.

\textbf{Figure~\ref{fig:motivation}a} shows a representative low-conflict mini-batch in which most cosine similarities are non-negative, indicating the group gradients are largely aligned. In contrast, \textbf{Figure~\ref{fig:motivation}b} presents a high-conflict mini-batch containing several group-gradient pairs with negative cosine similarities, showing that conflicts can arise within the same mini-batch. \textbf{Figure~\ref{fig:motivation}c} further summarizes the pairwise cosine similarities across all sampled mini-batches: among $563$ group-gradient pairs, $262$ pairs have negative cosine similarities, accounting for $46.5\%$ of the total. \textbf{Figure~\ref{fig:motivation}d} compares the validation $\Delta\mathrm{NLL}$ of mini-batches with low, medium, and high $C_{\mathrm{rate}}$, where the high-conflict subset exhibits a relatively lower median and contains negative updates, suggesting that severe gradient conflicts can be associated with less effective policy updates.

These findings suggest that group gradients induced by different queries within the same mini-batch can exhibit conflicts. Therefore, this leads to the central scientific question of this work: \textbf{When group gradients induced by different queries conflict within the same GRPO mini-batch, how can we obtain a reliable aggregated update direction?}

\subsubsection{Motivation Analysis}
\label{sec:motivation}
When group gradients point in conflicting directions, directly averaging them may not yield a reliable update direction.
Here, a reliable update direction refers to one that reasonably reflects the overall optimization tendency supported by the mini-batch, rather than being dominated by a few conflicting group gradients.
Such a direction should preserve the optimization information jointly provided by the mini-batch while limiting the disproportionate influence of particular conflicting signals.

Gradient uncertainty provides a possible basis for assessing such reliability of different group gradients during aggregation
\cite{ghavamzadeh2016bayesian,akella2021deep,balles2018dissecting}.
A group gradient can be viewed as a random variable represented by a probability distribution, which describes not only the possible values of the group gradient, but also how likely these values are.
When most of the probability mass is concentrated within a narrow region around the mean, the distribution mainly supports gradient values that are similar to one another.
The limited variation among these values corresponds to lower gradient uncertainty and higher reliability.
Conversely, when the probability mass is spread over a broader region, the distribution supports a wider range of substantially different gradient values.
The larger variation among these values corresponds to higher gradient uncertainty and lower reliability.
This notion of reliability characterizes how consistently a group-gradient direction is represented by its distribution, rather than whether that direction is necessarily correct.
\textbf{Figure~\ref{fig:conflict_illustration}} illustrates this perspective through a two-dimensional example.
The arrows $\mathbf{g}_1$ and $\mathbf{g}_2$ denote the mean gradients of two query groups, while the surrounding contours depict their probability distributions.
These two mean gradients point in conflicting directions, and their distributions exhibit different degrees of concentration.
The compact contours around $\mathbf{g}_1$ correspond to lower uncertainty and higher reliability, whereas the broader contours around $\mathbf{g}_2$ correspond to higher uncertainty and lower reliability.
Direct aggregation assigns the two group gradients equal coefficients without accounting for this difference in reliability.
Consequently, the more uncertain group gradient may exert a disproportionate influence on the directly aggregated direction $\mathbf{g}_{\mathrm{da}}$.

Motivated by the above, a natural way to address this problem is to account for group-gradient uncertainty during aggregation, placing greater emphasis on more reliable group gradients while limiting the influence of less reliable ones to obtain a more reliable update direction under conflicts.

\section{Methodology}
\label{sec:methodology}
Based on the above discussion, we propose Gradient Uncertainty-aware Policy Optimization (GUPO), a framework that incorporates the uncertainty of group gradients into the gradient aggregation of GRPO. Specifically, we first construct a probability distribution over the policy parameters under a Bayesian formulation and then estimate the induced distribution of each group gradient. We then transform the precision of the resulting gradient distributions into gradient evidence, from which the dimension-wise belief and group-level uncertainty are derived. Finally, we use the resulting uncertainty to calibrate the aggregation of group gradients within each mini-batch, assigning greater influence to lower-uncertainty gradients and attenuating the influence of higher-uncertainty ones. The overall framework is illustrated in \textbf{Figure~\ref{fig:overview}}, with the pseudo-code provided in the \textbf{Appendix}.

\subsection{Bayesian-based Group Gradient Distribution}
\label{sec:bayes_gradient_distribution}
In this subsection,  we formulate each query-level group gradient as a random variable by propagating parameter uncertainty to the corresponding gradient. Specifically, we construct a parameter distribution and use Monte Carlo sampling to estimate the induced distribution of each group gradient.

\begin{table*}[t]
\centering
\renewcommand{\arraystretch}{1.05}
\setlength{\tabcolsep}{4.2pt}
\resizebox{\textwidth}{!}{
\begin{tabular}{@{}llccccccc@{}}
\toprule
\textbf{Model} & \textbf{Method} & \textbf{GSM8K} & \textbf{MATH500} & \textbf{MinervaMATH} & \textbf{AMC 2023} & \textbf{AIME 2024} & \textbf{AIME 2025} & \textbf{Average Results} \\
\midrule
\multirow{9}{*}{\shortstack[l]{\textbf{DeepScaleR-1.5B}\\\textbf{-Preview}}}
& Vanilla & 89.6 & 85.2 & 24.6 & 83.0 & 42.8 & 36.7 & 60.3 \\
& GRPO \cite{shao2024deepseekmath} & 89.4 ($\downarrow$0.2) & 84.9 ($\downarrow$0.3) & 24.7 ($\uparrow$0.1) & 81.5 ($\downarrow$1.5) & 44.5 ($\uparrow$1.7) & 39.3 ($\uparrow$2.6) & 60.7 ($\uparrow$0.4) \\
& Length penalty \cite{arora2026training} & 88.5 ($\downarrow$1.1) & 83.2 ($\downarrow$2.0) & 23.0 ($\downarrow$1.6) & 77.3 ($\downarrow$5.7) & 40.3 ($\downarrow$2.5) & 30.3 ($\downarrow$6.4) & 57.1 ($\downarrow$3.2) \\
& ReST-MCTS \cite{zhang2024rest} & 89.9 ($\uparrow$0.3) & 84.8 ($\downarrow$0.4) & 23.9 ($\downarrow$0.7) & 83.4 ($\uparrow$0.4) & 45.5 ($\uparrow$2.7) & 39.5 ($\uparrow$2.8) & 61.2 ($\uparrow$0.9) \\
& GVPO \cite{zhang2026gvpo} & 90.4 ($\uparrow$0.8) & 85.7 ($\uparrow$0.5) & 25.3 ($\uparrow$0.7) & 83.6 ($\uparrow$0.6) & 46.1 ($\uparrow$3.3) & 39.7 ($\uparrow$3.0) & 61.8 ($\uparrow$1.5) \\
& Dr.GRPO \cite{liu2025understanding} & 90.0 ($\uparrow$0.4) & 85.3 ($\uparrow$0.1) & 25.1 ($\uparrow$0.5) & 82.1 ($\downarrow$0.9) & 45.8 ($\uparrow$3.0) & 39.6 ($\uparrow$2.9) & 61.3 ($\uparrow$1.0) \\
& GCPO \cite{gu2026group} & 90.5 ($\uparrow$0.9) & 86.3 ($\uparrow$1.1) & 25.9 ($\uparrow$1.3) & 84.1 ($\uparrow$1.1) & 46.7 ($\uparrow$3.9) & \textbf{40.3 ($\uparrow$3.6)} & 62.3 ($\uparrow$2.0) \\
& MRT \cite{qu2025optimizing} & 89.9 ($\uparrow$0.3) & 85.1 ($\downarrow$0.1) & 24.2 ($\downarrow$0.4) & 83.1 ($\uparrow$0.1) & 47.2 ($\uparrow$4.4) & 39.7 ($\uparrow$3.0) & 61.5 ($\uparrow$1.2) \\
& \textbf{GUPO (Ours)} & \textbf{91.7 ($\uparrow$2.1)} & \textbf{87.9 ($\uparrow$2.7)} & \textbf{26.4 ($\uparrow$1.8)} & \textbf{85.5 ($\uparrow$2.5)} & \textbf{48.7 ($\uparrow$5.9)} & 40.2 ($\uparrow$3.5) & \textbf{63.4 ($\uparrow$3.1)} \\
\midrule
\multirow{9}{*}{\shortstack[l]{\textbf{DeepSeek-R1-Distill}\\\textbf{Qwen-1.5B}}}
& Vanilla & 83.4 & 80.1 & 19.8 & 69.9 & 28.7 & 26.0 & 51.3 \\
& GRPO \cite{shao2024deepseekmath} & 84.5 ($\uparrow$1.1) & 80.3 ($\uparrow$0.2) & 22.1 ($\uparrow$2.3) & 70.5 ($\uparrow$0.6) & 29.8 ($\uparrow$1.1) & 27.3 ($\uparrow$1.3) & 52.4 ($\uparrow$1.1) \\
& Length penalty \cite{arora2026training} & 82.6 ($\downarrow$0.8) & 77.1 ($\downarrow$3.0) & 18.8 ($\downarrow$1.0) & 64.4 ($\downarrow$5.5) & 27.5 ($\downarrow$1.2) & 22.6 ($\downarrow$3.4) & 48.8 ($\downarrow$2.5) \\
& ReST-MCTS \cite{zhang2024rest} & 84.8 ($\uparrow$1.4) & 80.4 ($\uparrow$0.3) & 20.3 ($\uparrow$0.5) & 71.1 ($\uparrow$1.2) & 30.5 ($\uparrow$1.8) & 28.6 ($\uparrow$2.6) & 52.6 ($\uparrow$1.3) \\
& GVPO \cite{zhang2026gvpo} & 85.0 ($\uparrow$1.6) & 80.5 ($\uparrow$0.4) & 23.1 ($\uparrow$3.3) & 71.5 ($\uparrow$1.6) & 30.6 ($\uparrow$1.9) & 28.2 ($\uparrow$2.2) & 53.2 ($\uparrow$1.8) \\
& Dr.GRPO \cite{liu2025understanding} & 85.0 ($\uparrow$1.6) & 80.8 ($\uparrow$0.7) & 22.9 ($\uparrow$3.1) & 71.3 ($\uparrow$1.4) & 30.4 ($\uparrow$1.7) & 28.4 ($\uparrow$2.4) & 53.1 ($\uparrow$1.8) \\
& GCPO \cite{gu2026group} & 85.3 ($\uparrow$1.9) & 81.6 ($\uparrow$1.5) & 23.4 ($\uparrow$3.6) & 71.8 ($\uparrow$1.9) & 31.0 ($\uparrow$2.3) & 29.0 ($\uparrow$3.0) & 53.7 ($\uparrow$2.4) \\
& MRT \cite{qu2025optimizing} & 84.7 ($\uparrow$1.3) & 80.4 ($\uparrow$0.3) & 22.5 ($\uparrow$2.7) & 72.9 ($\uparrow$3.0) & 30.3 ($\uparrow$1.6) & 29.3 ($\uparrow$3.3) & 53.4 ($\uparrow$2.0) \\
& \textbf{GUPO (Ours)} & \textbf{86.3 ($\uparrow$2.9)} & \textbf{84.9 ($\uparrow$4.8)} & \textbf{24.8 ($\uparrow$5.0)} & \textbf{73.5 ($\uparrow$3.6)} & \textbf{33.2 ($\uparrow$4.5)} & \textbf{29.9 ($\uparrow$3.9)} & \textbf{55.4 ($\uparrow$4.1)} \\
\midrule
\multirow{9}{*}{\shortstack[l]{\textbf{DeepSeek-R1-Distill}\\\textbf{Qwen-7B}}}
& Vanilla & 91.6 & 87.4 & 42.1 & 85.1 & 55.5 & 50.2 & 68.6 \\
& GRPO \cite{shao2024deepseekmath} & 92.1 ($\uparrow$0.5) & 87.7 ($\uparrow$0.3) & 43.5 ($\uparrow$1.4) & 85.5 ($\uparrow$0.4) & 56.9 ($\uparrow$1.4) & 51.7 ($\uparrow$1.5) & 69.6 ($\uparrow$0.9) \\
& Length penalty \cite{arora2026training} & 91.1 ($\downarrow$0.5) & 83.7 ($\downarrow$3.7) & 39.5 ($\downarrow$2.6) & 81.2 ($\downarrow$3.9) & 53.8 ($\downarrow$1.7) & 46.9 ($\downarrow$3.3) & 66.0 ($\downarrow$2.6) \\
& ReST-MCTS \cite{zhang2024rest} & 92.0 ($\uparrow$0.4) & 87.9 ($\uparrow$0.5) & 42.8 ($\uparrow$0.7) & 85.7 ($\uparrow$0.6) & 57.1 ($\uparrow$1.6) & 52.4 ($\uparrow$2.2) & 69.7 ($\uparrow$1.0) \\
& GVPO \cite{zhang2026gvpo} & 92.9 ($\uparrow$1.3) & 88.5 ($\uparrow$1.1) & 44.2 ($\uparrow$2.1) & 86.3 ($\uparrow$1.2) & 57.5 ($\uparrow$2.0) & 52.1 ($\uparrow$1.9) & 70.3 ($\uparrow$1.6) \\
& Dr.GRPO \cite{liu2025understanding} & 92.3 ($\uparrow$0.7) & 88.2 ($\uparrow$0.8) & 44.0 ($\uparrow$1.9) & 86.4 ($\uparrow$1.3) & 57.4 ($\uparrow$1.9) & 52.3 ($\uparrow$2.1) & 70.1 ($\uparrow$1.5) \\
& GCPO \cite{gu2026group} & 92.6 ($\uparrow$1.0) & 89.1 ($\uparrow$1.7) & 45.0 ($\uparrow$2.9) & 87.3 ($\uparrow$2.2) & 58.3 ($\uparrow$2.8) & 53.0 ($\uparrow$2.8) & 70.9 ($\uparrow$2.2) \\
& MRT \cite{qu2025optimizing} & 92.2 ($\uparrow$0.6) & 88.4 ($\uparrow$1.0) & 44.3 ($\uparrow$2.2) & 86.0 ($\uparrow$0.9) & 57.0 ($\uparrow$1.5) & 52.4 ($\uparrow$2.2) & 70.1 ($\uparrow$1.4) \\
& \textbf{GUPO (Ours)} & \textbf{93.3 ($\uparrow$1.7)} & \textbf{89.5 ($\uparrow$2.1)} & \textbf{45.4 ($\uparrow$3.3)} & \textbf{87.8 ($\uparrow$2.7)} & \textbf{58.6 ($\uparrow$3.1)} & \textbf{53.9 ($\uparrow$3.7)} & \textbf{71.4 ($\uparrow$2.8)} \\
\bottomrule
\end{tabular}}
\caption{Pass@1 performance on mathematical reasoning datasets. We compare base models trained with different approaches. The best results under each base model are highlighted in bold. More results are provided in the \textbf{Appendix}.}
\label{tab:math_reasoning_results}
\end{table*}

\begin{figure*}[t]
    \centering
    \begin{subfigure}[t]{0.51\columnwidth}
        \centering
        \includegraphics[width=\linewidth]{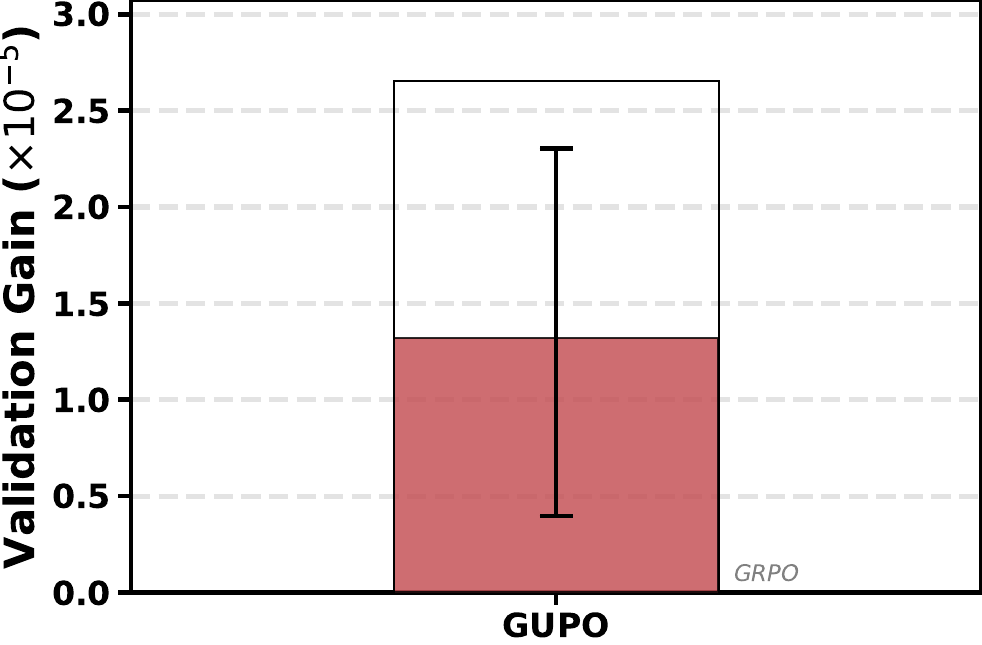}
        \caption{Mean Performance Gain}
        \label{fig:effect_mean}
    \end{subfigure}
    \hfill
    \begin{subfigure}[t]{0.48\columnwidth}
        \centering
        \includegraphics[width=\linewidth]{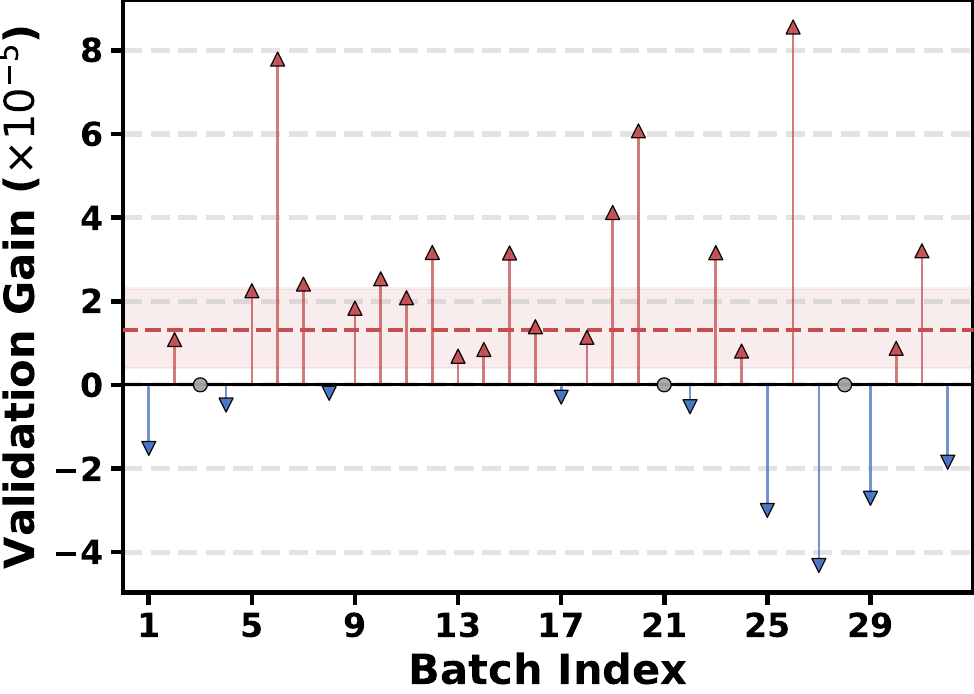}
        \caption{Batch Performance Gain}
        \label{fig:effect_batch}
    \end{subfigure}
    \hfill
    \begin{subfigure}[t]{0.52\columnwidth}
        \centering
        \includegraphics[width=\linewidth]{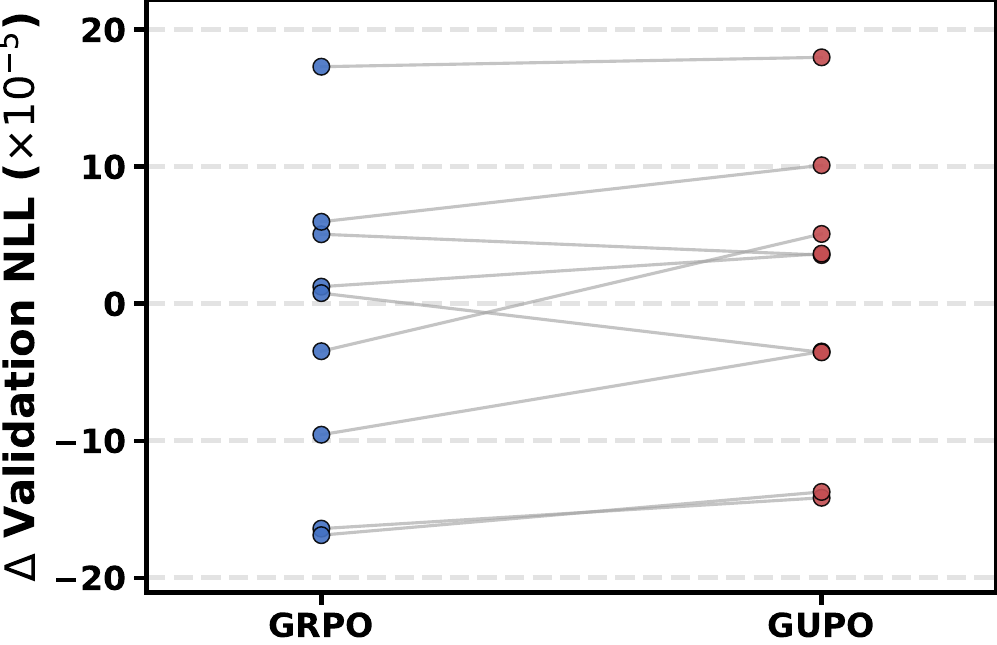}
        \caption{High-Conflict Batches}
        \label{fig:effect_pair}
    \end{subfigure}
    \hfill
    \begin{subfigure}[t]{0.55\columnwidth}
        \centering
        \includegraphics[width=\linewidth]{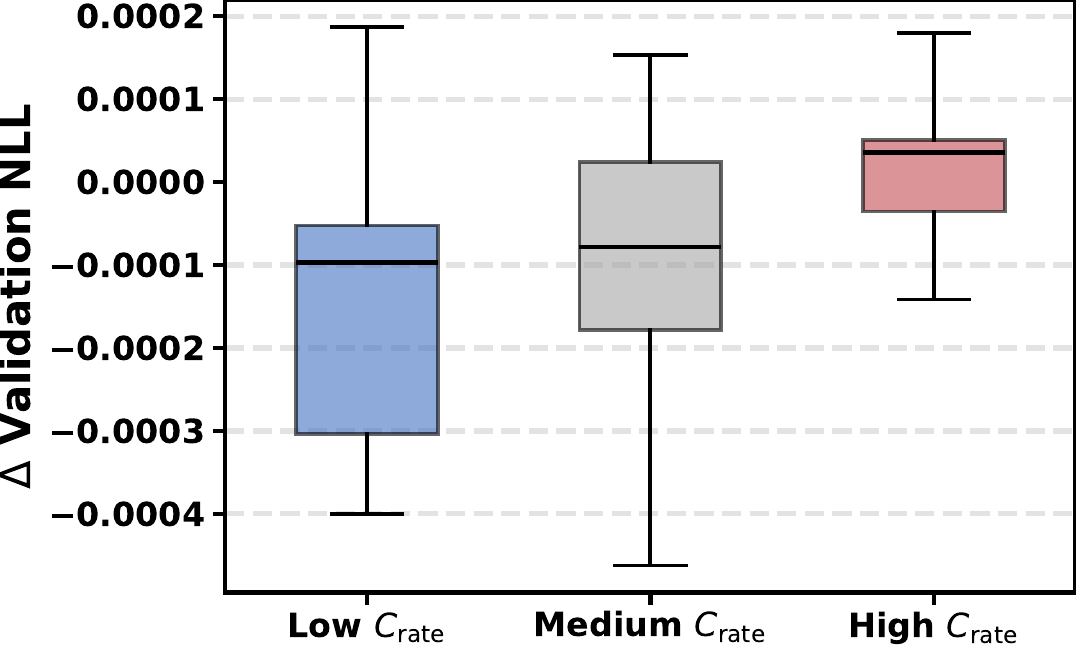}
        \caption{Conflict v.s. Performance}
        \label{fig:effect_bin}
    \end{subfigure}
    \caption{Effect of GUPO. (a) and (b) report the mean and batch-wise validation gains over GRPO. (c) shows the performance comparison on high-conflict mini-batches. (d) compares validation $\Delta\mathrm{NLL}$ produced by GUPO across different conflict levels.}
    \label{fig:gupo_effect}
\end{figure*}

\subsubsection{Posterior Distribution Estimation of Policy Parameters}
\label{sec:posterior_parameter}
Consider an optimization mini-batch containing $B$ queries: $\mathcal{B}=\left\{(q_b,\mathcal{O}_b)\right\}_{b=1}^{B}$, where $\mathcal{O}_b=\{o_{b,1},o_{b,2},\ldots,o_{b,G}\}$ contains the $G$ responses sampled for query $q_b$ from the old policy $\pi_{\theta_{\mathrm{old}}}$. Each response $o_{b,i}=\{o_{b,i,t}\}_{t=1}^{T_{b,i}}$ receives an outcome reward $r_{b,i}$. Its group-relative advantage is defined as $A_{b,i}=\frac{r_{b,i}-\bar{r}_b}{\sigma_{r,b}}$, where $\bar{r}_b$ and $\sigma_{r,b}$ denote the mean and standard deviation of the rewards within the $b$-th response group. We use $\mathcal{D}=\left\{(q_b,\mathcal{O}_b)\right\}_{b=1}^{B}$ to denote the rollout sequences contained in the mini-batch.

Directly constructing a posterior distribution over all parameters of an LLM is computationally prohibitive. We therefore adopt a Bayesian formulation for the parameter set $\Phi\subseteq\theta$ of the final trainable layer of the policy model. Given the rollout data $\mathcal{D}$, the posterior distribution of $\Phi$ is $p(\Phi\mid\mathcal{D})\propto p(\mathcal{D}\mid\Phi)p(\Phi)$, where $p(\Phi)$ is the prior distribution and $p(\mathcal{D}\mid\Phi)$ is the likelihood of the sampled responses. The specific prior and likelihood are in the \textbf{Appendix}.

The exact posterior is generally intractable because the likelihood depends nonlinearly on the policy parameters. We therefore use its expansion around $\Phi_0$ to construct a Gaussian approximation. Let $\Delta\Phi=\Phi-\Phi_0$. The second-order Taylor expansion of the log-posterior is
\begin{equation}
\begin{aligned}
\scalebox{0.95}{$\log p(\Phi|\mathcal{D})\approx c+\nabla_{\Phi}\log p(\Phi|\mathcal{D})|_{\Phi=\Phi_0}^{\top}\Delta\Phi-\frac{1}{2}\Delta\Phi^{\top}H\Delta\Phi,$}
\end{aligned}
\label{eq:posterior_taylor_expansion}
\end{equation}
where $c$ contains the terms independent of $\Phi$, and $H=-\nabla_{\Phi}^{2}\log p(\Phi|\mathcal{D})|_{\Phi=\Phi_0}$ is the negative Hessian matrix of the log-posterior. 
We omit the first-order term and retain the quadratic term to construct the Gaussian approximation $q(\Phi\mid\mathcal{D})=\mathcal{N}(\Phi\mid\Phi_0,H^{-1})$ centered at the current parameters (see \textbf{Appendix} for further discussion).

Directly computing the full Hessian matrix is infeasible for LLMs. 
We therefore construct a tractable approximation $\widetilde{H}$ using empirical Fisher information.
Specifically, for each generated token $o_{b,i,t}$, we compute the gradient of its log-probability with respect to the trainable parameters: $\mathbf{s}_{b,i,t}=\nabla_{\Phi}\log\pi_{\Phi}(o_{b,i,t}| q_b,o_{b,i,<t})|_{\Phi=\Phi_0}$, which describes how the generation probability of the token changes with the policy parameters. The empirical Fisher matrix is then obtained:
\begin{equation}
F_{\Phi}=\frac{1}{N_{\mathrm{token}}}\sum_{b=1}^{B}\sum_{i=1}^{G}\sum_{t=1}^{T_{b,i}}s_{b,i,t}s_{b,i,t}^{\top},
\label{eq:empirical_fisher}
\end{equation}
where $N_{\mathrm{token}}=\sum_{b=1}^{B}\sum_{i=1}^{G}T_{b,i}$ is the total number of tokens in the mini-batch.
The negative Hessian is approximated as $H\approx\widetilde{H}= \operatorname{Diag}(F_{\Phi})+\delta I,\;\delta>0$.
Accordingly, using this diagonal empirical Fisher approximation, we obtain
\begin{equation}
q(\Phi\mid\mathcal{D})=\mathcal{N}(\Phi\mid\Phi_0,\widetilde{H}^{-1}).
\label{eq:approximate_parameter_posterior}
\end{equation}

\subsubsection{Group Gradient Distribution Estimation}
\label{sec:gradient_distribution_estimate}
Based on the approximate parameter distribution, we next estimate the distribution induced over each group gradient.
As defined in \textbf{Problem Settings}, the group gradient associated with query $q_b$ is denoted by $\mathbf{g}_b(\theta)$. 
Because both the policy probabilities and their derivatives depend on the policy parameters, different values of $\Phi$ can produce different group gradients.
Therefore, when $\Phi\sim q(\Phi|\mathcal{D})$, the corresponding group gradient can be regarded as a random variable.

Since the distribution of $\mathbf{g}_b(\Phi)$ is generally difficult to derive in closed form, we approximate it through Monte Carlo sampling. Specifically, we draw $M$ parameter samples from the parameter distribution $\{\Phi^{(m)}\}_{m=1}^M\sim q(\Phi|\mathcal{D})$.
For each parameter sample $\Phi^{(m)}$, we construct the corresponding complete policy parameter set $\theta^{(m)}=\theta\left(\Phi^{(m)}\right)$, while keeping the remaining policy parameters fixed. We then compute the $b$-th group gradient $\mathbf{g}_b^{(m)}=\mathbf{g}_b(\theta^{(m)})$.
The mean and covariance of the $b$-th group gradient are defined as
\begin{equation}
\begin{aligned}
\boldsymbol{\mu}_b&=\frac{1}{M}\sum_{m=1}^{M}\mathbf{g}_b^{(m)},\\
\Sigma_b&=\frac{1}{M-1}\sum_{m=1}^{M}(\mathbf{g}_b^{(m)}-\boldsymbol{\mu}_b)(\mathbf{g}_b^{(m)}-\boldsymbol{\mu}_b)^{\top}.
\end{aligned}
\label{eq:group_gradient_covariance}
\end{equation}

To avoid constructing a full covariance matrix in the high-dimensional gradient space, we use the element-wise variance vector $\boldsymbol{\sigma}_b^2$ instead of the full covariance matrix.
The probability distribution of $\mathbf{g}_b$ is then approximated as
\begin{equation}
q(\mathbf{g}_b\mid\mathcal{D})\approx\mathcal{N}(\mathbf{g}_b\mid\boldsymbol{\mu}_b,\operatorname{Diag}(\boldsymbol{\sigma}_b^2)).
\label{eq:group_gradient_distribution}
\end{equation}

\subsection{Gradient Uncertainty-aware Policy Optimization}
\label{sec:gupo_method}
In this subsection, we incorporate gradient uncertainty into the aggregation of group gradients. Specifically, we first derive the gradient precision and transform it into gradient evidence. The evidence is then used to derive dimension-wise belief and group-level uncertainty. The resulting group-level uncertainty is subsequently used to calibrate the aggregation of group gradients within the same mini-batch.

For the $b$-th group, let $\sigma_{b,d}^{2}$ denote the variance of its gradient in the $d$-th dimension. We first compute the corresponding gradient precision $\lambda_{b,d}=\frac{1}{\sigma_{b,d}^{2}}$, which describes the concentration of the gradient distribution in that dimension. 
A larger $\lambda_{b,d}$ corresponds to a more concentrated gradient distribution in the corresponding dimension, indicating that the gradient varies less across the sampled parameters.

We then map the gradient precision to non-negative gradient evidence $e_{b,d}=\phi\left(\lambda_{b,d}\right)=\lambda_{b,d}^{\,s}$, where $s>0$ controls the sensitivity of the evidence to the gradient precision. A larger gradient precision produces stronger evidence, while a smaller gradient precision produces weaker evidence.

Inspired by subjective logic and evidence theory \cite{jsang2018subjective,sensoy2018evidential}, the corresponding Dirichlet concentration parameter is defined as $\alpha_{b,d}=e_{b,d}+1$. 
The dimension-wise belief and group-level uncertainty are then defined as
\begin{equation}
b_{b,d}=\frac{e_{b,d}}{S_b},\quad u_b=\frac{K}{S_b},\quad S_b=\sum_{d=1}^{K}\alpha_{b,d}.
\label{eq:gradient_belief_uncertainty}
\end{equation}
where $S_b$ is the Dirichlet strength and $K$ denotes the total number of gradient dimensions. 
$b_{b,d}$ represents the belief supported by the evidence in the $d$-th dimension, while $u_b$ summarizes the uncertainty of the $b$-th group gradient.

Then, we use the group-level uncertainty to adjust the aggregation of group gradients.
The uncertainty-aware weight of the $b$-th group is defined as $\omega_b=\frac{1-u_b}{\sum_{\ell=1}^{B}(1-u_{\ell})}$. To preserve the original aggregation structure of GRPO, we further combine $\omega_b$ with the original gradient coefficient $\frac{1}{B}$: $\widetilde{\omega}_b=\frac{1-\eta}{B}+\eta\omega_b$, where $\eta\in[0,1]$ controls the degree to which group-gradient uncertainty adjusts the original gradient aggregation (see the \textbf{Appendix} for more discussion). 
The final GUPO gradient is
\begin{equation}
\nabla_{\theta}\mathcal{J}_{\mathcal{B}}^{\mathrm{GUPO}}(\theta)=\sum_{b=1}^{B}\widetilde{\omega}_b\mathbf{g}_b(\theta).
\label{eq:gupo_gradient}
\end{equation}

Under this aggregation, group gradients with lower uncertainty receive greater contributions, whereas those with higher uncertainty receive smaller contributions. Consequently, this aggregation aims to provide a more reliable update direction for policy optimization.

\section{Experiments}
\label{sec:experiment}
In this section, we conduct experiments on various benchmarks to evaluate GUPO. More details, experiments, and analysis are provided in the \textbf{Appendix}.

\subsection{Experimental Settings}
\label{sec:exp_setting}

\subsubsection{Benchmarks and Baselines}
We compare our method with a series of representative post-training baselines, including GRPO~\cite{shao2024deepseekmath}, Length Penalty~\cite{arora2026training}, ReST-MCTS \cite{zhang2024rest},  GVPO~\cite{zhang2026gvpo}, Dr.~GRPO~\cite{liu2025understanding}, GCPO~\cite{gu2026group}, and MRT~\cite{qu2025optimizing}.
We evaluate our method on several widely used reasoning benchmarks: AIME 2024, AIME 2025, AMC 2023, MATH500~\cite{hendrycks2021measuring}, MinervaMATH~\cite{lewkowycz2022solving}, and GSM8K~\cite{cobbe2021training}. AIME and AMC evaluate competition-level mathematical reasoning. MATH500 and MinervaMATH cover diverse mathematical problem-solving tasks. GSM8K focuses on multi-step grade-school arithmetic reasoning. We report Pass@1 accuracy on each benchmark and the average performance across all six benchmarks.

\subsubsection{Implementation Details}
We conduct experiments on three base models at two scales: DeepScaleR-1.5B-Preview, DeepSeek-R1-Distill-Qwen-1.5B, and DeepSeek-R1-Distill-Qwen-7B. We use a learning rate of $1\times10^{-6}$, a weight decay of $0.01$, and a global batch size of $256$. All experiments are conducted on an NVIDIA H100 GPU cluster. More implementation details are provided in the \textbf{Appendix}.

\begin{figure}[t]
    \centering
    \begin{subfigure}[t]{0.49\columnwidth}
        \centering
        \includegraphics[width=\linewidth]{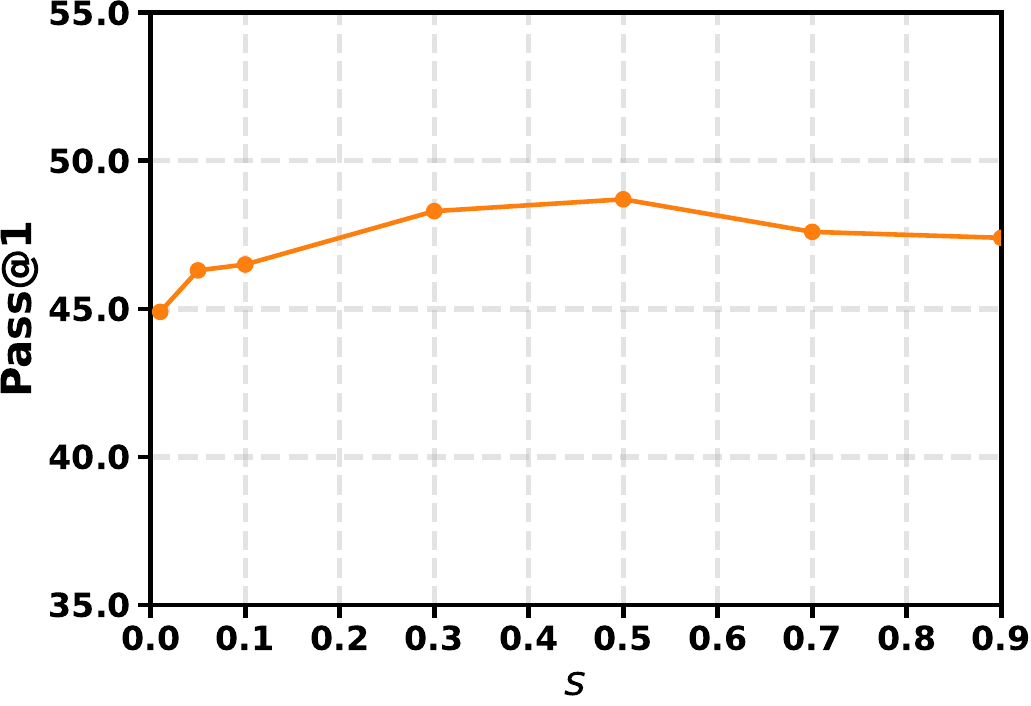}
        \caption{Parameter $s$}
        \label{fig:abla_s}
    \end{subfigure}
    \hfill
    \begin{subfigure}[t]{0.49\columnwidth}
        \centering
        \includegraphics[width=\linewidth]{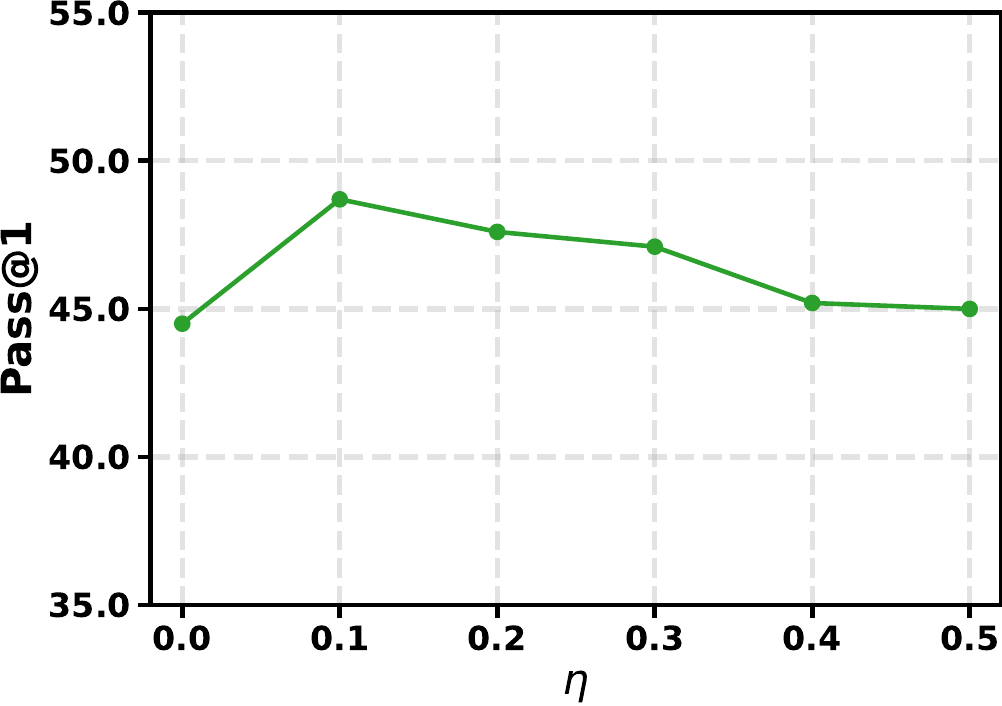}
        \caption{Parameter $\eta$}
        \label{fig:abla_eta}
    \end{subfigure}
    \caption{Parameter sensitivity.}
    \label{fig:exp_abla}
\end{figure}

\subsection{Main Results}
\label{sec:exp_results}
\subsubsection{Performance Comparison}
\label{sec:main_exp}
We compare GUPO with representative baselines across multiple benchmarks and base models. As shown in \textbf{Table \ref{tab:math_reasoning_results}}, the results demonstrate the effectiveness of GUPO, which incorporates group-gradient uncertainty into gradient aggregation.

\subsubsection{Ablation Study about Parameter Sensitivity}
\label{sec:abla}
We conduct analysis of evaluating evidence sensitivity $s$ at $[0.01, 0.05, 0.1, 0.3, 0.5, 0.7, 0.9]$, and aggregation coefficient $\eta$ over $[0, 0.5]$ with an interval of $0.1$. \textbf{Figure~\ref{fig:exp_abla}} shows the best performance is achieved at $s=0.5$ and $\eta=0.1$, which are our final setting. More results are in the \textbf{Appendix}.

\subsubsection{Analysis}
\label{sec:exp_analysis}
We further analyze the effect of GUPO under group-gradient conflicts. As shown in \textbf{Figure~\ref{fig:gupo_effect}}, GUPO can improve validation performance over GRPO across sampled mini-batches, with the paired comparison on high-conflict mini-batches providing further evidence. The results suggest that GUPO can mitigate the degradation in update effectiveness associated with severe group-gradient conflicts.

\section{Conclusion}
\label{sec:conclusion}
In this work, we investigate conflicts among group gradients induced by different queries within the same GRPO mini-batch and show that such conflicts can be associated with less effective policy updates. To obtain a reliable update direction, we propose Gradient Uncertainty-aware Policy Optimization (GUPO), which models each group gradient as a probability distribution under a Bayesian formulation, estimates its uncertainty, and incorporates the resulting uncertainty into gradient aggregation. By enhancing the contribution of lower-uncertainty group gradients and mitigating the influence of highly uncertain ones, GUPO can enable the aggregated gradient to better reflect the optimization tendency supported by the mini-batch. Experiments across multiple reasoning benchmarks demonstrate the effectiveness of GUPO.

\appendix

\bibliography{aaai2027}


\end{document}